\documentclass[letterpaper,twocolumn,10pt]{article}
\usepackage{usenix}

\usepackage{tikz}
\usepackage{amsmath}
\usepackage{amsthm}
\usepackage{amssymb}
\usepackage[utf8]{inputenc}
\usepackage[T1]{fontenc}
\usepackage{xspace}
\usepackage[most]{tcolorbox}
\usepackage{tabularx}
\usepackage{subcaption}
\usepackage{stfloats}
\usepackage{enumitem}
\usepackage{multirow}
\usepackage{nicefrac}
\usepackage[most]{tcolorbox}
\usepackage{booktabs}
\usepackage{thmtools}
\usepackage{thm-restate}
\usepackage{balance}
\usepackage{float}
\restylefloat{table}

\newcommand{\hs}{H\&S\xspace}
\newcommand{\qa}{QA\xspace}
\newcommand{\repliqa}{\texttt{RepliQA}\xspace}
\newcommand{\bioasq}{\texttt{BioASQ}\xspace}
\newcommand{\llmjudge}{LLM-as-a-Judge\xspace}
\newcommand{\llmjudges}{LLM-as-Judges\xspace}
\newcommand{\deberta}{DeBERTaV3\xspace}
\newcommand{\squad}{\texttt{SQuAD2}\xspace}
\newcommand{\hl}[1]{\setlength{\fboxsep}{2pt}\colorbox{yellow}{#1}}
\newcommand{\minwords}{\texttt{min\_words}\xspace}
\newcommand{\llm}{\ensuremath{\mathcal{L}\xspace}}
\newcommand{\llminput}{\ensuremath{\mathbb{P}\xspace}}
\newcommand{\llmoutput}{\ensuremath{\mathbb{O}\xspace}}

\newtheorem{definition}{Definition}[section]

\definecolor{myorange}{RGB}{240, 140, 0}

\begin{document}

\date{}

\title{\Large \bf Highlight \& Summarize: RAG without the jailbreaks}

\author{
{\rm Giovanni Cherubin}\\
Microsoft Security Response Center
\and
{\rm Andrew Paverd}\\
Microsoft Security Response Center
}

\maketitle

\begin{abstract}
Preventing jailbreaking and model hijacking of Large Language Models (LLMs) is an important yet challenging task.
When interacting with a chatbot, malicious users can input specially crafted prompts that cause the LLM to generate undesirable content or perform a different task from its intended purpose.
Existing systems attempt to mitigate this by hardening the LLM's system prompt or using additional classifiers to detect undesirable content or off-topic conversations.
However, these probabilistic approaches are relatively easy to bypass due to the very large space of possible inputs and undesirable outputs. 

We present and evaluate \emph{Highlight \& Summarize} (\hs), a new design pattern for retrieval-augmented generation (RAG) systems that prevents these attacks \emph{by design}.
The core idea is to perform the same task as a standard RAG pipeline (i.e., to provide natural language answers to questions, based on relevant sources) without ever revealing the user's question to the generative LLM.
This is achieved by splitting the pipeline into two components: a highlighter, which takes the user's question and extracts (``highlights'') relevant passages from the retrieved documents, and a summarizer, which takes the highlighted passages and summarizes them into a cohesive answer.
We describe and implement several possible instantiations of \hs and evaluate their  responses in terms of correctness, relevance, and quality.  
For certain question-answering (\qa) tasks, the responses produced by \hs are judged to be as good, if not \emph{better}, than those of a standard RAG pipeline.
\end{abstract}

\section{Introduction}
Retrieval-augmented Generation (RAG) is proving to be a sound and reliable solution for answering questions using documents from a knowledge base.
From customer support systems to search engines, RAG combines the ability of LLMs to answer questions expressed in natural language with the efficiency of vector databases for retrieving information from large data repositories, to provide a robust production-ready solution for many applications.
The core idea behind RAG is that, when a user asks a question, the system first retrieves a set of relevant documents from the knowledge base and passes these to a generative LLM together with the user's question; on this basis, the LLM produces an answer to the question, which is returned to the user.

There are, alas, many ways an adversary can exploit RAG systems, depending on which inputs the adversary can control. %
In this paper, we focus on the scenario where the knowledge base is trusted, while the users' inputs are untrusted and potentially malicious.
A real-world example is a chatbot that is deployed by a company to answer questions based on the company's curated knowledge base (e.g., consisting of official FAQs or policy documents).

In this setting, a malicious user could attack the system by inputting specially crafted prompts to achieve various objectives.
Firstly, they could try to \emph{jailbreak} the system to make the LLM generate offensive content that harms the reputation of the company.
A more subtle jailbreak could be to trick the LLM into generating unintended outputs that misrepresent the company's intent, for example, persuading the chatbot to offer discounts on the company's products.
In some cases, the chatbot's output may constitute a legally binding statement from the company.\footnote{It was recently reported that Air Canada had to pay compensation to a customer for misleading information provided by their chatbot. \url{https://www.theguardian.com/world/2024/feb/16/air-canada-chatbot-lawsuit}.}
Alternatively, the malicious user could repurpose the generative LLM to perform some other task (e.g., asking questions unrelated to the company) -- this is sometimes referred to as \emph{model hijacking}~\cite{operationBizarreBazaar,fuzzyAI}.

\begin{figure*}
    \centering
    \includegraphics[width=0.85\textwidth]{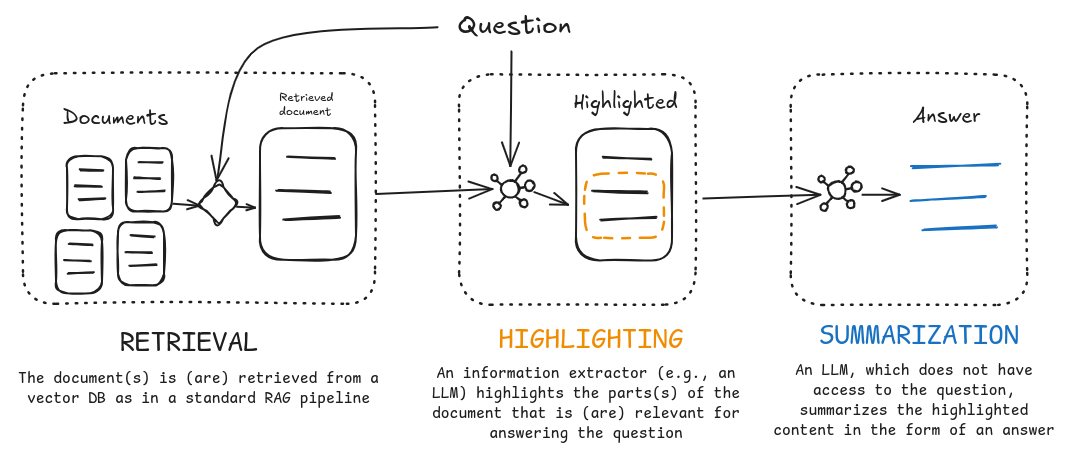}
    \caption{HS-enhanced RAG pipeline. A malicious user asking a question to this system cannot jailbreak the LLM, by design.
    }
    \label{fig:hs}
\end{figure*}

In this paper, we introduce and evaluate \emph{Highlight \& Summarize} (\hs), a new design pattern for RAG systems that prevents these attacks by design.
Our approach consists of two components ~(\autoref{fig:hs}):
First, the \emph{highlighter} component inspects the retrieved documents and extracts passages from them that are relevant to answering the user's question -- the digital equivalent of \emph{highlighting} those passages.
The highlighter can be instantiated using existing methods for extractive \qa, or using modern generative LLMs.
Secondly, the \emph{summarizer} component, which is an LLM, generates a coherent answer by summarizing the highlighted passages.
\textbf{Crucially, the summarizer never sees the user's question}.
This mitigates the above attacks whilst maintaining the utility of the RAG system on its intended \qa task.

We formalize the security guarantees through a theoretical analysis (with an accompanying \texttt{Lean} formalization) showing how \hs exponentially decreases the control that an attacker can exercise over the \qa pipeline.
We show empirically that \hs thwarts all non-adaptive attacks from a large prompt injection dataset as well as a strong novel \emph{adaptive highlighting} attack specifically designed to challenge this technique.

We implement several variants of the \hs design pattern and evaluate the accuracy of their responses on two widely-used question answering (\qa) datasets: \repliqa and \bioasq.
Using various LLM-as-a-Judge implementations, we compare the \hs generated responses against those of a standard ``vanilla'' RAG implementation, both in a direct head-to-head comparison as well as using independent ratings for the answer's correctness, relevance, and quality.
Our results show that certain \hs implementations produce answers that are comparable, and sometimes better, than those of a standard RAG implementation.
For example, the ``Two Steps'' \hs method achieves correctness scores that are equivalent (\repliqa: 95\% vs 94\%) or even slightly better (\bioasq: 89\% vs 86\%) than those of Vanilla RAG, with comparable performance on all other metrics.
In 1-to-1 Elo ranking, \hs is sometimes better than a standard RAG pipeline
(e.g., Elo on \repliqa: 1170 vs 1104, \bioasq: 1181 vs 1197).

We also perform several ablation studies on different components and aspects of \hs to provide deeper insights into \emph{why} our technique achieves these strong results.

In summary, we make the following contributions:
\begin{itemize}
    \item We present Highlight \& Summarize (\hs), a new design pattern for RAG systems that mitigates jailbreaks and model hijacking by design.
    \item We formalize the security guarantees of \hs, and we thoroughly evaluate it against both non-adaptive and adaptive attacks.
    \item We implement several different variants of \hs and show empirically that their generated outputs are comparable to those of a standard RAG pipeline.
\end{itemize}

To enable further analysis of \hs, without incurring the monetary and environmental costs of rerunning our experiments, we release both our code and experimental data, including the responses generated by the various RAG pipelines and the associated ratings given by the \llmjudges (see the Open Science appendix for further details).

\section{Background and related work}
\label{sec:background}

\paragraph{Retrieval-augmented generation (RAG).} At a high level, the intended interaction between as user and a typical LLM-based RAG application would be: 

1. The user submits a question via a chat-style interface.

2. The application searches its knowledge base and retrieves documents related to the user's question.

3. The retrieved documents are concatenated to the user's question and the combined text is input into an LLM.

4. The LLM generates a response, which the application returns to the user.

\noindent This type of RAG application is widely used, for example in customer support services.\footnote{\url{https://careersatdoordash.com/blog/large-language-modules-based-dasher-support-automation}.}\textsuperscript{,}\footnote{\url{https://medium.com/tr-labs-ml-engineering-blog/better-customer-support-using-retrieval-augmented-generation-rag-at-thomson-reuters-4d140a6044c3}.}
The benefits of RAG in such scenarios include the ability to ground the LLM's response on company-specific information, reduce hallucinations, and update the knowledge base without retraining the LLM.
However, as explained above, such systems are still vulnerable to jailbreaking and model hijacking by malicious users.

\paragraph{Preventing jailbreaks and model hijacking.} Several techniques have been proposed to detect or prevent jailbreaking of LLM-based systems, including those used in RAG.
The main classes of defenses include: finetuning the LLM to reduce the risk of undesirable output; using defensive system prompts to increase the difficulty of jailbreaking~\cite{Microsoft2025SafetySystemMessages,mu2025closerlook}; applying different types of classifiers to the inputs to detect potential jailbreaks~\cite{alon2023detectingjailbreaks, jain2023baseline, hu2024tokenlevel, pisano2024bergeron, OpenAIModeration}; and preprocessing the input to remove or reduce the impact of potential jailbreaks~\cite{rebedea2023nemo, jain2023baseline, robey2024smoothllm, kumar2025certifying}.
To mitigate model hijacking~\cite{operationBizarreBazaar,fuzzyAI} (i.e., repurposing the LLM for a different task), techniques such as NeMo~\cite{rebedea2023nemo} support programmable guardrails that allow the application owner to specify dialogue flows by canonicalizing inputs.
Our solution, \hs, takes a completely different approach by ensuring that the user's input is never input to the summarizer LLM.

\paragraph{System-level defenses.} Similarly to defenses such as CaMeL~\cite{debenedetti2025caml} and FIDES~\cite{costa2025fides}, \hs is a system-level defense in that it deterministically limits the attacker's capabilities.
However, \hs focuses on the opposite threat model:
whereas those defenses prevent malicious sources to affect benign users,
in \hs' threat model the retrieved data is trusted but the user inputs are not.
\hs can be seen as a realization of the Context-Minimization design pattern~\cite{beurerkellner2025designpatterns}.

\paragraph{Passage retrieval and extractive \qa.}
Passage retrieval and extractive \qa are well-established research fields in the domain of natural language processing (NLP).
Passage retrieval is the process of extracting from a set of documents one or more texts that are relevant to answering a question.
Several techniques have been developed, with applications in the medical and legal sectors~\cite{o1975retrieval,o1980answer,callan1994passage,jiang2006extraction}.
With a similar goal, extractive \qa aims to select (brief) portions of text from a larger document in order to answer a question.
Extractive \qa methods are largely based on modern NLP architectures, with BERT-like models being the most successful~\cite{wang2022survey,pearce2021comparative,lee2016learning}.
Passage retrieval and extractive \qa are typically combined in practical systems~\cite{jm3,prasad2023meetingqa}.
In an \hs pipeline, the highlighter has a very similar goal to extractive \qa, and
two of the instantiations of the highlighter we describe in \autoref{sec:hs} use extractive \qa models.

\section{Threat model and assumptions}
\label{sec:threat-model}

Some users of a RAG application may be adversarial. 
We assume a strong adversary who has full knowledge of all the documents in the knowledge base;
for example, this may be the case if the knowledge base consists of published documentation or FAQ web pages.
The adversary is able to submit arbitrary questions and observe the responses.
The adversary may want to achieve the following goals:
\begin{itemize}
    \item Repurpose the application to perform a different task. For example, using a customer service chatbot to summarize large amounts of text, at the expense of the application owner.
    \item Jailbreak the application and cause it to generate arbitrary output. This could include content that causes reputational damage to the application owner, or even constitutes an unintentional yet legally enforceable commitment from the application owner.
\end{itemize}

\noindent
The goal of the application owner is to prevent both of the above classes of attacks.
We assume the application owner has inspected all documents in the knowledge base and confirmed that they contain only trustworthy information.
This is a realistic assumption as the application owner has full control of the knowledge base and can apply arbitrary preprocessing and filtering.
Note that we do not aim to defend against other types of attacks on RAG systems, such as extracting information~\cite{jiang2024ragthief, peng2025dataextraction} or poisoning the knowledge base~\cite{zou2024poisonedrag, zhang2025corruptrag, xiang2024certifiably}.

\section{Highlight \& Summarize}
\label{sec:hs}

We propose \hs, a new design pattern for RAG systems to mitigate jailbreaking and model hijacking attacks.
The main design principle is that a (malicious) user must be unable to provide arbitrary inputs to the LLM that generates the response.
We achieve this by separating the generative \qa process into two steps (\autoref{fig:hs}):

\begin{itemize}
    \item \textbf{Highlighting:} The highlighter component takes the retrieved documents and selects (``highlights'') text passages from these documents that are relevant to answering the user's question.
    \item \textbf{Summarization:} The summarizer component takes only the text selected by the highlighter and summarizes it in the form of a coherent answer to \emph{some} question, which is returned to the user. The user's question is never shown to the summarizer.
\end{itemize}

The following security property is enforced in \hs:

\begin{tcolorbox}[colback=red!5!white,colframe=red!75!black]
  For a security parameter \minwords, each of the text passages
    returned by
    the highlighter in a \hs pipeline must be a contiguous
    match of at least \minwords words from the trusted documents;
    furthermore, text extracts must be non-overlapping.
\end{tcolorbox}

\noindent

This property can be deterministically enforced by the pipeline, by checking that every output of the highlighter contains
contiguous and non-overlapping exact matches from the retrieved documents,
and that they contain
at least \minwords words each.

By virtue of this design, malicious users are unable to directly influence the outputs of the system -- which are provided by the summarizer.
In \autoref{sec:security}, we support this claim by considering an important class of
adaptive attacks and how they are countered.

\paragraph{Implementing \hs.}
Since \hs is a new \emph{design pattern}, it can be implemented in various different ways.
In particular, there are many ways to implement the highlighter component, ranging from extractive QA models through to prompt-tuned LLMs.
In \autoref{sec:hs:highlighter}, we describe five different implementations of the highlighter.
In contrast, the summarizer is relatively straight-forward, so we present a single implementation in \autoref{sec:hs:summarizer}. 
We refer to any combination of a highlighter and summarizer as an \textit{\hs pipeline}.

\subsection{Highlighter implementations}
\label{sec:hs:highlighter}

We describe five different highlighter implementations that have been chosen to provide a broad overview of the design space. They differ in terms of computational overhead and \qa capabilities.

\paragraph{\hs Baseline.}
This highlighter is a zero-shot prompt-tuned LLM that is tasked with extracting relevant information from the retrieved documents.
Since the output of the highlighter LLM might deviate slightly from the exact text in the retrieved documents (e.g., due to the model's internal randomness), we use fuzzy string matching between the LLM's output and the documents to identify the precise text from the documents to be highlighted.
We use \texttt{RapidFuzz}\footnote{\url{https://github.com/rapidfuzz/RapidFuzz}.} with a threshold of 95 in our experiments.

\paragraph{\hs Structured.}
This highlighter improves upon \hs Baseline by asking the highlighter LLM to first output an answer to the user's question and then to highlight the relevant text from the retrieved documents.
We again use Azure OpenAI's structured output to enforce the following format:\\
\texttt{\{``answer'': str, ``text\_extracts'': list[str]\}}.

\noindent
The generated answer is not passed to the summarizer.
We observed that asking the highlighter LLM to first produce this output helped with grounding its responses, thereby producing better highlights from the text.
As with \hs Baseline, we again use fuzzy string matching to ensure that the highlighted text is taken directly from the retrieved documents.

\paragraph{\hs Two Steps.}
This highlighter enforces the Chain Of Thought implied in \hs Structured by employing
two separate LLM calls. In the first call, we have an LLM answer the question;
in the second call, the LLM is given the question, the answer, and the documents,
and it is asked to extract the text needed to answer the question.
As with all other highlighter implementations, we ensure that Two Step returns
verbatim contiguous text from the original document and nothing else.

\paragraph{\hs Span.}
This highlighter only returns the start and end of the passages to highlight.
This has two benefits. First, it considerably reduces the costs of the pipeline, since the highlighter LLM outputs considerably less tokens
than the other two (\autoref{fig:hs-costs}).
Secondly, it simplifies the task of matching the highlighted text against the document: because only a small part of the document is output,
the LLM is less likely to introduce typos; for this reason, we did not find it necessary to implement fuzzy matching for the outputs of \hs Span
(the pipeline simply rejected any non-matching texts).

\paragraph{\hs \deberta.}
This highlighter is an extractive \qa model from the BERT family.
In our experiments, we use \deberta, which uses disentangled attention and a better mask decoder to improve upon BERT and RoBERTa.
We use two versions of this model:
\textit{\hs \deberta (\squad).} is a \deberta model that was fine-tuned\footnote{\url{https://huggingface.co/deepset/deberta-v3-base-squad2}.} on the \squad dataset~\cite{rajpurkar-etal-2016-squad}.
Since this model was fine-tuned for short-span extractive \qa, as encouraged by the \squad dataset, its performance in our experiments was lacking: the highlighter of an \hs pipeline
is best served by a tool that outputs longer span passages to a question.
For this reason, we also employ \textit{\hs \deberta (\repliqa)}, which is a \deberta model that we fine-tuned on the \repliqa dataset (see \autoref{sec:setup}).
We train the model to return the gold passage of an answer, rather than the expected answer, which significantly improves the performance of the \hs pipeline.
Fine-tuning this model on splits 0-2 of the \repliqa dataset (\autoref{sec:setup}) took around 7 hours on an NVIDIA A100 GPU.

\subsection{Summarizer implementation}
\label{sec:hs:summarizer}

We evaluate one summarizer that is common across all \hs implementations.
The summarizer is a zero-shot prompt-tuned LLM that is tasked with i) guessing what question the highlighted text was intended to answer, and ii) reformulating the highlighted text in form of an answer.
Only the answer is returned to the user.
The guessed question is not returned to the user -- it's purpose is to ground the generated answer, and to aid in evaluation.
We use Azure OpenAI's structured output option\footnote{\url{https://learn.microsoft.com/en-us/azure/ai-services/openai/how-to/structured-outputs}.}, which forces the LLM to give a response matching a desired format (in this case, \texttt{\{``guessed\_question'': str, ``answer'': str}\}); we found this helps with the quality of the responses, and allows us to evaluate what question the LLM guessed.
Unless otherwise specified, we employ OpenAI's \texttt{GPT-4.1 mini}~\cite{achiam2023gpt} as our default generative LLM.

\section{Security analysis}
\label{sec:security}

Fundamentally, all forms of jailbreaking and model hijacking attacks involve some type of adversarial input to an LLM, which causes the LLM to generate undesirable outputs or perform an unintended task. 
In the widely-used setting of a RAG-powered system with a trusted knowledge base, the key idea of \hs is to ensure that the adversary cannot provide \emph{direct} inputs to the generative LLM that produces the final output (i.e., the summarizer);
consequently, this design should prevent an attacker from controlling
arbitrary outputs.

In this section,
we first show theoretically that H\&S exponentially decreases the control that an attacker
can exercise over the \qa pipeline.

Then, we show how it thwarts all non-adaptive attacks from a large prompt injections dataset (\autoref{sec:security:non_adaptive}).
Finally, and most importantly, we carefully look into ways \textit{adaptive} attackers
might evade the security of H\&S, by influencing indirectly the inputs of the generative LLM through the highlighted text, and we show how enforcing a
\minwords security parameter mitigates the vast majority of them (\autoref{sec:security:adversarial_highlighting}).

\subsection{Theoretical analysis}

The security of a \hs pipeline comes from the fact that the inputs to the generative LLM (\textit{summarizer}) are limited to being exact contiguous extracts of trusted text documents.
This significantly reduces the ways an attacker can control the LLM to make it misbehave.
We provide a theoretical foundation to support this intuition.
We formulate our analysis upon the definition of \textit{control region of an LLM}, which is novel to the best of our knowledge, and show that \hs exponentially decreases an attacker's control of the LLM's outputs.

\subsubsection*{\hs prevents arbitrary control.}

Formally, an autoregressive generative LLM is a randomized algorithm $\llm: \llminput \mapsto \Delta(\llmoutput)$ that maps input prompts (i.e., sequences of tokens) to a probability distribution over output strings, where $\Delta(\llmoutput)$ is the space of distributions on $\llmoutput$.
We use the shortcut notation $P(\llm(s) = o)$ to mean $P_{O \sim \llm(s)}(O = o)$, where $O$ is a random variable with distribution $\llm(s)$.

We define the control region of an LLM as the set of outputs that can be obtained with at least probability $\beta >0$.

\begin{definition}[LLM Control Region]
    For a parameter $\beta \in (0, 1]$, the  control region of an LLM $\mathcal{L}$
    is:
    $$\mathcal{C}_{\beta}(\llm) = \{o \in \llmoutput \mid  \exists p\in\llminput: P(\llm(p) = o) \geq \beta\}$$
\end{definition}

Intuitively, this captures the set of strings that a user can reliably obtain as outputs from the LLM (i.e., control with non-negligible probability) by providing inputs from $\llminput$.
For example, a user can have an LLM output the string $o=$ ``\texttt{This is an example}'' with non-zero probability (in fact, with high probability with modern instruction-tuned LLMs) by prompting the LLM with ``\texttt{Output the following string verbatim and nothing else: \textit{This is an example}}''.
Hence, the sentence $o$ belongs to the control region for some $\beta >0$.

A very informative result shows that, even for probabilistic LLMs, the number of possible outputs achievable with non-negligible probability is bounded by the size of their inputs:

\begin{restatable}[Bounded control region]{lemma}{bounded}
\label{thm:bounded-control-region}
    The size of the control region for an LLM is bounded by the size of the
    input space:
    $$|\mathcal{C}_{\beta}(\llm)| \leq \frac{|\llminput|}{\beta} \,.$$
\end{restatable}

\textit{Proofs are deferred to the appendix, which also includes a reference to \texttt{Lean} formalizations and proofs of these results.}

The above shows that the number of outputs that one can control strictly depends on the cardinality of the input set of an LLM, regardless of the randomness in the algorithm.
Because an \hs pipeline reduces the LLM's input space (hence, the output space), from the set of all input tokens $\llminput$ to the set of contiguous strings, we can informally say that:
\begin{tcolorbox}[colback=blue!5!white,colframe=blue!75!black]
An \hs pipeline prevents the attacker from having \textit{arbitrary} control over the outputs.
That is, it restricts the outputs that they can obtain from the LLM.
\end{tcolorbox}

\subsubsection*{Exponential descrease in control.}

We now quantify by how much \hs decreases the attacker's control.
Suppose the LLM's input space is the set of strings of at most
$K$ tokens, $\llminput = \Sigma^{\leq K}$,
where $\Sigma$ is the set of tokens and $K$ is the LLM's token limit.
Let us now consider the summarizer LLM of an H\&S pipeline.
For a retrieved document $D \in \Sigma^*$, the highlighting step of H\&S is a map $h_D(\llminput)$ that returns a contiguous string from the document $D$.\footnote{More precisely, the space is restricted to strings with length between \minwords and $K$; we disregard this here for simplicity, as it has no bearing on the theoretical conclusions. Similarly, we could have multiple text extracts, but as long as their number is constant it does not make any difference.}
An H\&S pipeline reduces the control region from
$\mathcal{C}_{\beta}(\llm)$ to
$\mathcal{C}_{\beta}(\llm \circ h_D)$.
Precisely, we have:

\begin{restatable}[Exponential decrease in control]{theorem}{exponential}
\label{thm:exponential}
Consider an LLM $\llm: \Sigma^K \mapsto \Sigma^*$
and an H\&S pipeline $\llm \circ h_D$ operating on a document of
length $N=|D|$.
Assume there are constants $\alpha, \beta \in (0, 1]$ s.t.:
$|\mathcal{C}_{\beta}(\llm)| \geq \alpha |\llminput|$.
Then
$$
\frac{|\mathcal{C}_{\beta}(\llm \circ h_D)|}{|\mathcal{C}_{\beta}(\llm)|}
    = O(K \, N \,|\Sigma|^{-K})
$$
\end{restatable}

This results makes a basic coverage assumption on the LLM:
it assumes that the LLM
is expressive enough that some non-zero proportion $\alpha>0$ of its potential outputs
can be obtained with non-zero probability.
For example, consider a variation of the input prompt discussed before,
$p(x) =$ ``\texttt{Output the following string verbatim and nothing else:} $x$'',
where $x$ is a string.
We can expect a modern LLM to output $x$ with probability $\beta >0$
for \emph{some proportion} of strings $x \in \llmoutput$.\footnote{The assumption is required to hold for an arbitrary $\alpha > 0$ proportion of outputs, which can be small.
One could potentially lift this assumption and derive similar results from the
way softmax scores are derived for LLMs: if (as it is desirable) all softmax scores
are required to be $< 1$, one may show that a similar exponential bound to \autoref{thm:exponential} can be obtained.
}
This assumption aligns with the LLM's effectiveness:
an LLM developer would like to have a large $\beta$ control over a large portion of the
output space, since this implies a well-behaving LLM.

In summary, \autoref{thm:exponential} confirms that:

\begin{tcolorbox}[colback=blue!5!white,colframe=blue!75!black]
An \hs pipeline reduces the control region of the summarizer LLM \textit{exponentially}
in $K$, the token limit of the LLM.
That is, it exponentially reduces the number of outputs that an attacker can control.
\end{tcolorbox}

\subsection{Non-adaptive attacks}
\label{sec:security:non_adaptive}

Evaluating LLM jailbreaks requires an automated method for detecting whether the jailbreak succeeded.
Inspired by the recent LLMail-Inject challenge~\cite{abdelnabi2025llmail}, we use \textit{tool calling} as a well-defined attack target.
We give the LLM the ability to call a (simulated) email sending tool by outputting a specific string (e.g., \texttt{send\_email()}).
For example, a customer service chatbot may have this type of tool to send an email to the customer support team if a user asks about certain topics, but the users should not be able to control when the LLM triggers this tool call.
The attacker's goal is therefore to cause the LLM to trigger this tool call.\footnote{This is very similar to the setting of a recent vulnerability found by Zenity Labs: \url{https://labs.zenity.io/p/a-copilot-studio-story-2-when-aijacking-leads-to-full-data-exfiltration-bc4a}.
}
For purposes of evaluation, we give \textit{both} LLMs (i.e., highlighter and summarizer) the ability to trigger this tool call.
However, in practice only the summarizer would have this ability, so a successful attack against \hs would need to trick the summarizer into making this tool call.

For these experiments, we use 1,028 successful attack prompts from the LLMail-Inject challenge dataset (Scenario 2)~\cite{abdelnabi2025llmail}.
Our setup differs slightly from that of the challenge:
in the challenge, the adversarial inputs were retrieved from the user's emails, whereas in our setting we input these directly as the user's prompt.
However, this modified setup does not affect the results: successful prompts from LLMail-Inject are usually successful in our setup.

\begin{table}
    \caption{Evaluation on the LLMail-Inject dataset (Scenario 2), showing the percentage of all jailbreaks that succeeded in calling the prohibited tool and the percentage of all jailbreaks that succeeded in calling the tool with valid arguments.
    }
    \label{tab:llmail}
    \begin{tabularx}{1.02\columnwidth}{Xrr}
    \toprule
     & Tool called & Arguments valid \\
    \midrule
    RAG & 81\% & 53\% \\
    H\&S (Highlighter only) & 93\% & 63\% \\
    H\&S & 0\% & 0\% \\
    \bottomrule
    \end{tabularx}
\end{table}

\autoref{tab:llmail} shows the percentage of all inputs for which the tool was called (with any arguments) or called with valid arguments. 
As expected, the vast majority of the attack inputs succeed in jailbreaking the RAG pipeline, but none are successful against \hs.
The results on \hs (Highlighter only) confirm that our highlighter LLM, like any other, is susceptible to jailbreaks when processing adversarial input.
This emphasizes the need for system-level defenses, such as \hs.

\subsection{Adaptive highlighting attack}
\label{sec:security:adversarial_highlighting}

As explained in \autoref{sec:threat-model}, we assume a strong adversary who may have full knowledge of \hs and the contents of the knowledge base used by the system under attack.
We envision that such an attacker could attempt to subvert \hs with the following adaptive attack, which is somewhat reminiscent of traditional system attacks such as Return-Oriented Programming (ROP)~\cite{shacham2007geometry}.
For a target sentence, the attacker
i) browses the RAG knowledge base to find a document that contains (most of) the words in the target sentence, 
ii) causes that document to be retrieved, and
iii) asks the \hs highlighter to only highlight the desired words.
For example, as shown in \autoref{fig:adaptive_adversarial}, if the attacker wants the system to output ``You won a \$10 voucher'', they could try to persuade the highlighter to highlight only specific words from the retrieved document.

\begin{figure}
    \centering
    \begin{tcolorbox}[colback=white, colframe=myorange!75!white, 
      title=Reimbursement\_Guidelines.pdf, fonttitle=\bfseries, coltitle=white, 
      boxrule=0.8mm, arc=4mm, top=2mm, bottom=2mm, left=2mm, right=2mm]
    
    \hl{You} may submit a request for reimbursement if the expense falls within the approved categories outlined in Appendix D. Once the claim is verified and approved, it is recorded as a completed transaction and marked as “\hl{won}” in the internal tracking system. A confirmation email will be sent to the claimant within three (3) business days. If the reimbursement includes \hl{a \$10} threshold adjustment, the system will automatically generate a voucher code for accounting purposes. This \hl{voucher} is non-transferable and must be used within the same fiscal quarter.%
    \end{tcolorbox}
    \caption{Adaptive highlighting attack example.}
    \label{fig:adaptive_adversarial}
\end{figure}

While this \emph{adversarial highlighting} attack is theoretically possible, it is made extremely challenging thanks to our simple \hs design constraint: we only allow the highlighter to highlight non-overlapping, contiguous passages of at least a certain length \minwords.
For example, by simply setting $\minwords=5$, it is impossible for the attacker to highlight the sentence ``You won a a \$10 voucher'' in \autoref{fig:adaptive_adversarial},
which in turn makes the same attack virtually impossible to succeed.

Importantly, \hs makes it viable to \textit{check in advance}
whether the knowledge base documents contain any string that may lead to jailbreaks: as discussed in \autoref{sec:discussion}, one could
explore the documents looking for strings that may lead to jailbreaks.
Consequently, \hs transforms the very challenging problem of detecting any possible jailbreak in the LLM's inputs into the significantly simpler task of inspecting the knowledge base to check that it does not contain strings that can be used to trigger undesirable outputs.

\section{Experimental setup}
\label{sec:setup}

In the following sections, we evaluate the quality of the responses of the various \hs pipelines, and compare them with a vanilla \qa LLM in a standard RAG setting.
Evaluating LLMs in \qa tasks is notoriously challenging~\cite{chen2019evaluating,adlakha2024evaluating,chiang2023can,zheng2023judging,verga2024replacing,oosterhuis2024reliable,zheng2023judging}.
We select datasets that prevent confounding factors (e.g., the fact that the LLM may not use the context document for answering), and we employ many evaluation metrics, ranging from standard to novel ones, to provide a comprehensive overview of the performance of \hs pipelines.

\begin{table*}
\caption{Metrics used for evaluating individual \hs components and the full \hs pipeline.}
\label{tab:metrics}
\begin{tabularx}{\textwidth}{lcXc}
    \toprule
    Name & Type & Description & Ref \\
    \midrule
    Recall & Token & Proportion of tokens
in the reference answer are present in the model's response. & \cite{adlakha2024evaluating}\\
    K-Precision & Token & Proportion of tokens in the model's reponse
that are present in the gold passage. & \cite{adlakha2024evaluating}\\
    Poll Multihop Correctness & LLM & Correctness of a generated response against a reference answer using few-shot learning. & \cite{verga2024replacing} \\
    Reliable CI Relevance & LLM & Relevance of a passage to a query based on a four-point scale: Irrelevant, Related, Highly relevant, Perfectly relevant. & \cite{oosterhuis2024reliable} \\
    MTBench Chat Bot Response Quality & LLM & Quality of the response based on helpfulness, relevance, accuracy, depth, creativity, and level of detail, assigning a numerical grade. & \cite{zheng2023judging} \\
    ComparisonJudge & LLM & Compare two answers to the same question and select either a winner, a tie, or a tie where neither answer is acceptable. & Ours \\
    \bottomrule
\end{tabularx}
\end{table*}

\subsection{Datasets}
When evaluating a generative LLM for RAG pipelines,
the LLM should ground its answers on the documents provided as context,
and \textit{not on any of its training data}.
For example, if a RAG application is based on policy documents, any answer it gives must
be coming strictly from those documents and not from other policies the LLM may have
memorized in training.

\paragraph{\repliqa.}
To mitigate this confounding factor, 
we conduct our main experiments using the \repliqa dataset~\cite{monteiro2024repliqa}.
This is a human-created dataset that contains questions based on natural-looking documents about \emph{fictitious} events or people.
By doing so, we ensure that the performance is not affected by the ability of LLMs to memorize their training data.

The \repliqa dataset consists of 5 splits (numbered from 0 to 4), which were released gradually over a year.
Each split contains 17,955 examples.
In our evaluation, we used the most recent split (\texttt{split\_3}, which was released on April 14th, 2025). This ensures that the LLM we use was not trained on this data.
During data analysis, we observed 10 mislabeled instances in this split, which we corrected manually.

Each entry in the \repliqa dataset contains a \emph{gold passage}, called \texttt{long\_answer}, which is a substring of the retrieved document, selected by
a human annotator, that is relevant to answering the question.
We use this field as part of our performance measurements, as well as for fine-tuning (on a separate split of the dataset) a \deberta extraction \qa model to implement the \hs \deberta (\repliqa) pipeline.

\paragraph{\bioasq.}
We also include experiments based on the \texttt{rag-mini-bioasq} dataset, henceforth \bioasq, which is a subset of the training dataset that was used for the BioASQ Challenge~\cite{bioasq}.
This dataset contains biomedical questions, as well as answers and documents.
Since we do not use this dataset for training, we use both the training and test splits for evaluation, which gives a total of 4,719 samples.

\subsection{Evaluation metrics}

We compare each \hs pipeline's answer with the expected answer from the dataset using several metrics.
We evaluate the quality of the responses based on three types of evaluation metrics, as summarized in \autoref{tab:metrics}.
We report the costs of H\&S
in terms of time-to-respond and
token counts in \autoref{sec:costs}.

\paragraph{Pairwise comparison judge.}
We use a zero-shot prompt-tuned LLM with the task of deciding a ``winner'' between two alternative answers to a question.
This judge also has the option to declare a ``tie'' between the answers, or to decide that neither answer is acceptable.
In our implementation, we randomize the order of the two answers to mitigate any potential ordering bias.

\paragraph{\llmjudge.}
We use prompt-tuned LLMs that rate the responses with respect to various criteria~\cite{chiang2023can, zheng2023judging}.
Specifically, we adopt three standard \llmjudge implementations, so as to provide a diverse judgment~\cite{verga2024replacing,oosterhuis2024reliable,zheng2023judging}.

\paragraph{Token-based metrics.}
We use quantitative comparisons between the tokens of the expected answer (or gold passage) and the given answer.
Specifically, we use token-based metrics to evaluate the correctness (``Recall'') and faithfulness (``K-Precision''),
based on the work by Adlakha et al.~\cite{adlakha2024evaluating}.

\section{Evaluating the full \hs pipeline}
\label{sec:e2e-comparison}

We first evaluate the full \hs pipelines in comparison with a standard (``vanilla'') RAG system. 
Since \hs only modifies RAG generation step, we hold the retrieval step constant for each comparison; that is, we compare the pipelines on their ability to answer questions based on the same set of retrieved documents.
We compare the pipelines using the ComparisonJudge and the \llmjudges, as well as on their ability to decline to answer when no answer can be provided.

\subsection{Pairwise comparisons}

\begin{table}
\caption{Direct pairwise comparison of all pipelines by the ComparisonJudge LLM. Win rate uses standard Elo scoring (win=1, tie=0.5, loss=0).}
\label{tab:average_wins}
\begin{tabularx}{1.02\columnwidth}{p{1mm}Xrr}
\toprule
& & Elo Score & Wins\\
\midrule
\multirow{7}{*}{\rotatebox[origin=c]{90}{\repliqa}}
& H\&S Span Highlighter & 1170 & 60\% \\
& H\&S Baseline & 1145 & 62\% \\
& H\&S Structured Highlighter & 1105 & 63\% \\
& Vanilla RAG & 1104 & 61\% \\
& H\&S Two Steps Highlighter & 1098 & 67\% \\
& H\&S BERT Extractor (RepliQA) & 702 & 22\% \\
& H\&S BERT Extractor (SQuAD2) & 673 & 14\% \\
\midrule
\multirow{7}{*}{\rotatebox[origin=c]{90}{\bioasq}}
& Vanilla RAG & 1197 & 66\% \\
& H\&S Structured Highlighter & 1181 & 70\% \\
& H\&S Two Steps Highlighter & 1104 & 70\% \\
& H\&S Baseline & 1099 & 66\% \\
& H\&S Span Highlighter & 854 & 31\% \\
& H\&S BERT Extractor (RepliQA) & 799 & 28\% \\
& H\&S BERT Extractor (SQuAD2) & 762 & 19\% \\
\bottomrule
\end{tabularx}
\end{table}

We use the ComparisonJudge to perform direct pairwise comparisons between all pipelines, including standard RAG.
For each question in our evaluation set, we compare all pairs of pipeline outputs, resulting in one of three outcomes: a clear preference for one response, a tie (both equally good), or neither (both inadequate).

\begin{figure}
    \centering
    \includegraphics[width=0.9\linewidth]{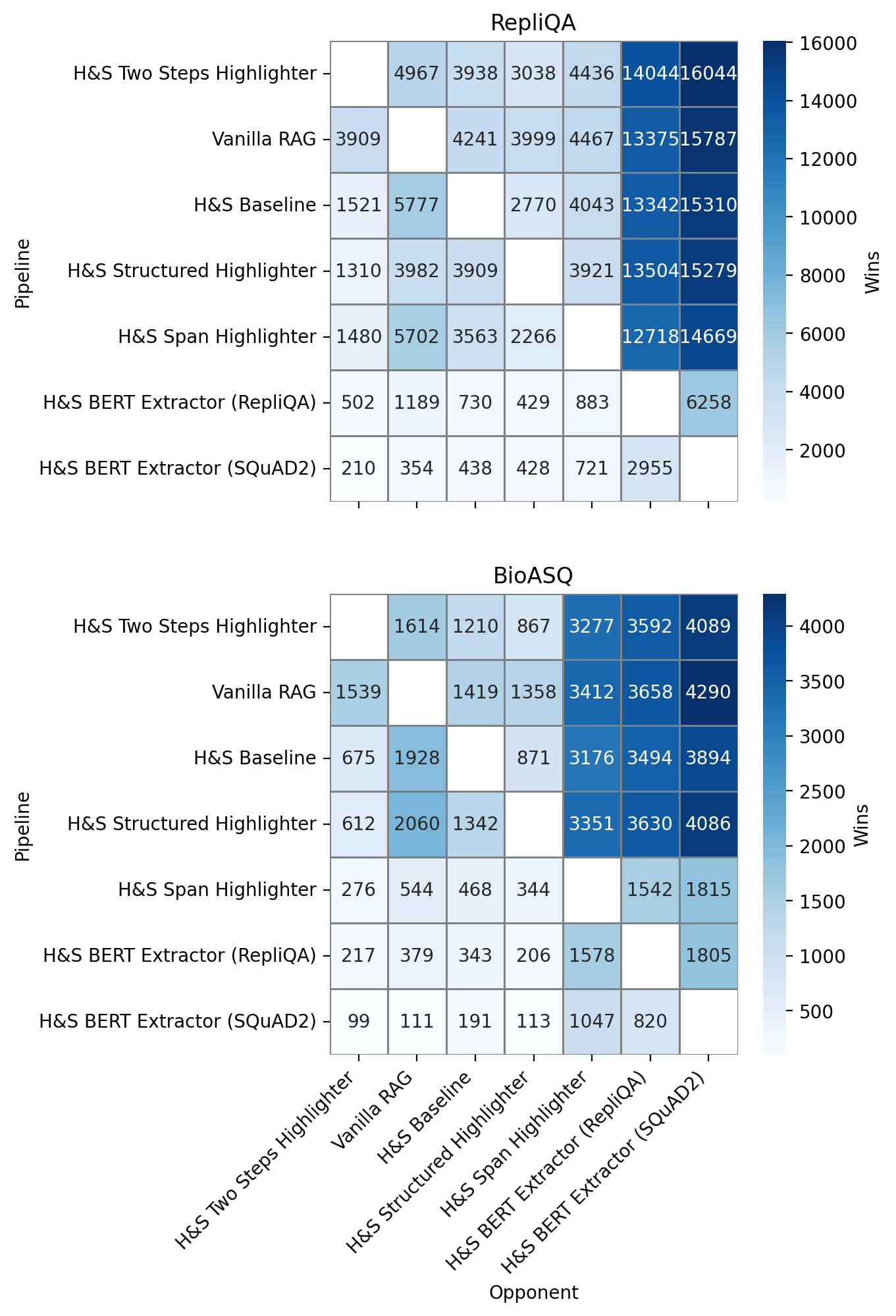}
    \caption{Wins of each pipeline in the direct pairwise comparison by the ComparisonJudge. Ties are omitted.}
    \label{fig:wins-heatmap}
\end{figure}

We compute Elo ratings using pairwise comparisons.
The Elo score, invented in the context of chess for ranking players based on 1-to-1 matches~\cite{elo1967proposed}, has recently become a popular metric for comparing LLMs (e.g., by Chiang et al.~\cite{chiang2024chatbot}).
We use the Bradley-Terry model to compute Elo scores from these pairwise outcomes, treating ties as half a win for each model (the standard convention).
We also report wins rate:
${\rm Wins Rate} = \frac{{\rm Wins} + 0.5 \times {\rm Ties}}{{\rm Total Games}}$.
Elo rankings are in \autoref{tab:average_wins}, whereas the full pairwise results are in \autoref{fig:wins-heatmap} and \autoref{tab:battle_results}.

\autoref{fig:wins-heatmap} shows that, as expected, vanilla (unsecured) RAG has the largest number of wins when discarding ties.
However, all \hs methods are competitive, particularly \hs Two Steps which even has more ``wins'' on the \repliqa dataset.
When looking at Elo scoring (\autoref{tab:average_wins}), which accounts for ties, we observe that \hs Two Steps outperforms the vanilla RAG baseline for both datasets.
Interestingly, most LLM-based \hs methods perform comparably to (or better than) the baseline in terms of Elo ranking.

\textbf{This is an exciting result because it suggests that, for these particular tasks, \hs simultaneously gives security against malicious users as well as \emph{comparable} (sometimes \emph{better}) response accuracy than standard RAG.}
We attribute this success to the fact that \hs forces, by design, a form of chain-of-thought reasoning, by prompting the model to answer the question in (at least) two steps.
In \autoref{sec:discussion}, we discuss cases where we expect \hs to work well, and applications where it may not work as well.

\subsection{Correctness, relevance, and quality}

\begin{figure}
    \centering
    \includegraphics[width=0.85\linewidth]{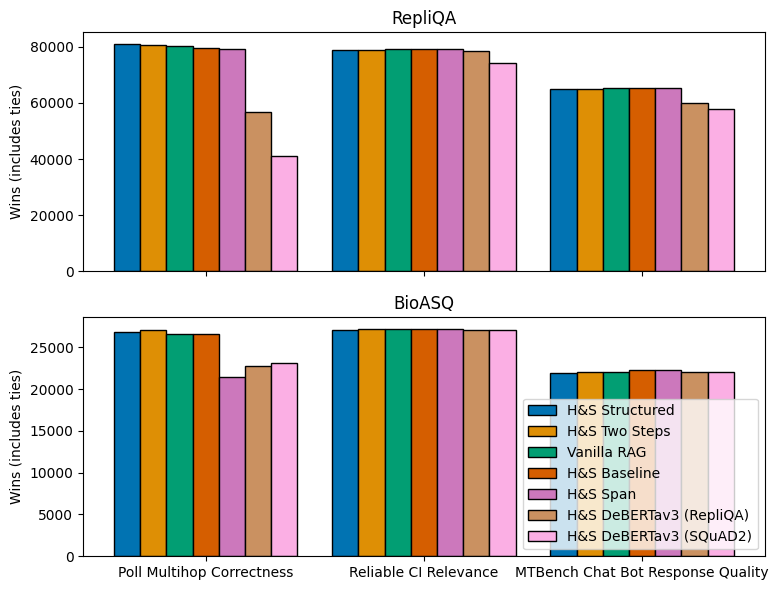}
    \caption{Number of times each pipeline had the highest rating, including ties, according to \llmjudges.}
    \label{fig:graders-wins}
\end{figure}

\begin{figure*}
    \centering

    \includegraphics[width=\linewidth]{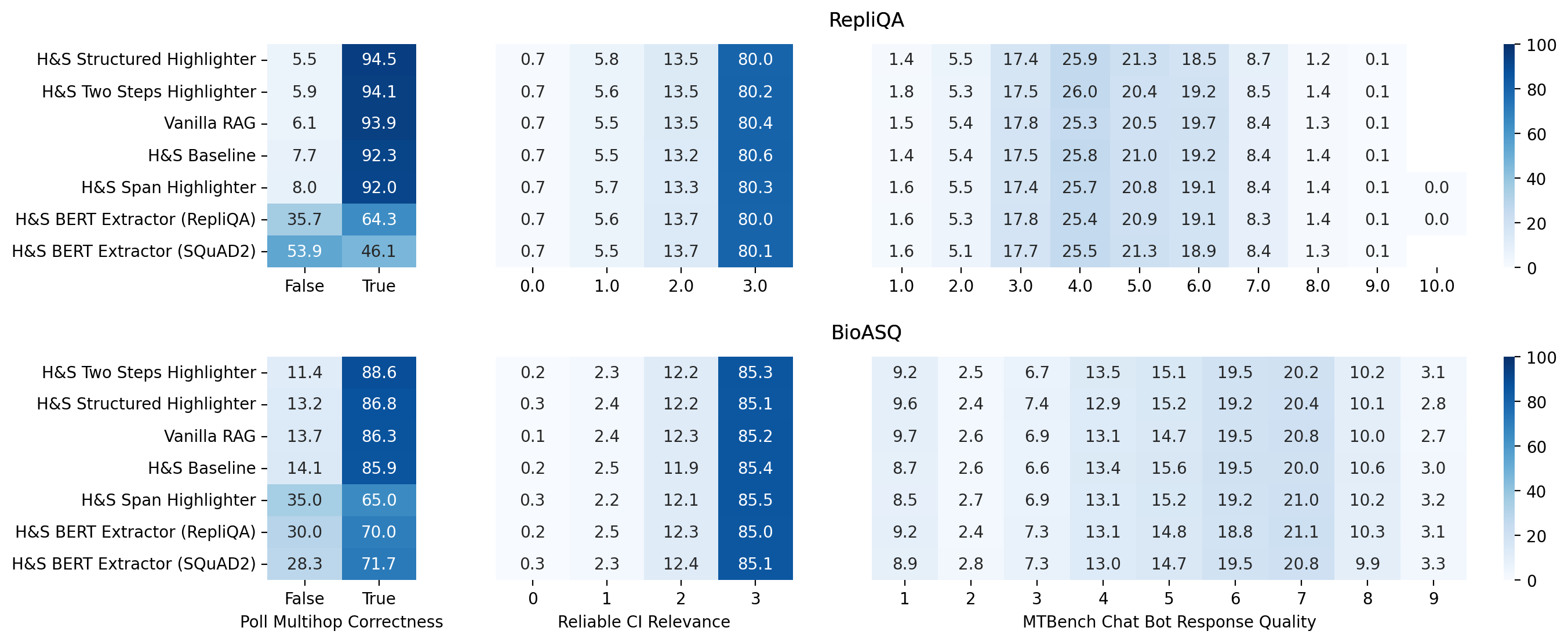}
    \caption{Response's evaluation via \llmjudges on the two datasets, measuring: correctness, relevance, and quality.}
    \label{fig:llmjudges}
\end{figure*}

We now evaluate the pipelines independently via the use of three \llmjudges, which independently evaluate correctness, relevance, and quality (as detailed in \autoref{tab:metrics}).

\autoref{fig:graders-wins} reports summary statistics on the number of times each pipeline received the highest rating out of all the pipelines (including tied highest).
We observe that, whereas the pipelines are very similar in terms of relevance and quality of the responses, there are notable differences in correctness.

On \repliqa, the LLM-based highlighters (\hs Structured and \hs Two Steps) achieve the highest correctness scores, with vanilla RAG close behind. The BERT-based extractors show notably lower correctness, with the \squad-trained model performing worst; this is likely because it was not trained on this domain.
On BioASQ, a similar pattern emerges: \hs Two Steps and \hs Structured lead on correctness, while \hs Span underperforms.

These results suggest that the correctness metric is the most discriminative, while relevance and quality judges show little differentiation between pipelines. The strong performance of LLM-based highlighters on correctness indicates that accurate span selection is crucial for answer quality, whereas the extractive BERT approaches—particularly when applied out-of-domain—struggle to identify the most relevant passages.

\autoref{fig:llmjudges} shows the distribution of ratings from each judge.
We discuss each judge separately.

\paragraph{Correctness.}
Poll Multihop Correctness evaluates the correctness of a generated answer against the reference answer (binary: True/False).
On \repliqa, the LLM-based highlighters (\hs Structured and \hs Two Steps) achieve the highest correctness, with over 13,400 correct answers out of approximately 14,200 answerable questions.
We observe that \hs \deberta (\squad) performs poorly on \repliqa, achieving only 5,921 correct answers—likely due to its base highlighter being fine-tuned for returning short answers rather than extracting relevant passages.
In contrast, the same base model fine-tuned on a separate split of \repliqa achieves 8,676 correct answers, suggesting that fine-tuning on in-domain data substantially improves extractive QA performance.
On \bioasq, \hs Two Steps leads (27,092), while \hs Span underperforms (21,442). Notably, the \repliqa-tuned \deberta performs worse on \bioasq than the \squad-tuned variant, reinforcing that data distribution highly impacts performance and that more varied fine-tuning datasets should be used in practice.

\paragraph{Relevance.}

The Reliable CI Relevance metric captures how relevant an answer is to a question on a 0–3 scale.
We observe minimal differentiation across pipelines: all methods achieve near-perfect relevance, with the vast majority of answers receiving a score of 3.
On \repliqa, Vanilla RAG and \hs Baseline show marginally higher counts at the top rating, while on \bioasq, \hs Span edges slightly ahead. However, these differences are negligible in practice.

\paragraph{Response quality.}

MTBench Chat Bot Response Quality evaluates responses on a 1–10 scale.
We observe that ratings cluster around 5–6 for both datasets, with minimal variation across pipelines.
\hs Baseline and Vanilla RAG achieve marginally higher quality scores, while the BERT-based extractors lag slightly behind.
Based on manual inspection of the judge's explanations, we observe that the judge tends to demand additional information beyond what is available in the retrieved document.
This suggests the judge is better suited for rating LLMs on open-ended questions without source documents, rather than in RAG settings where answers are necessarily constrained by the retrieved context.

\subsection{K-Precision and Recall}

\begin{table}[h]
    \caption{K-precision and recall of the predicted answers against the reference passage and answer.}
    \label{tab:kprec-recall}
    \begin{tabularx}{1.02\columnwidth}{p{1mm}Xrr}
    \toprule
    & Pipeline & K-Precision & Recall \\
    \midrule
    \multirow{7}{*}{\rotatebox[origin=c]{90}{\repliqa}}
    & Vanilla RAG & 0.76 & 0.78 \\
    & H\&S Structured Highlighter & 0.70 & 0.65 \\
    & H\&S Span Highlighter & 0.69 & 0.64 \\
    & H\&S Baseline & 0.69 & 0.63 \\
    & H\&S Two Steps Highlighter & 0.69 & 0.64 \\
    & H\&S \deberta (\repliqa) & 0.63 & 0.39 \\
    & H\&S \deberta (\squad) & 0.44 & 0.22 \\
    \midrule
    \multirow{7}{*}{\rotatebox[origin=c]{90}{\bioasq}}
    & Vanilla RAG & 0.55 & 0.44 \\
    & H\&S Baseline & 0.37 & 0.42 \\
    & H\&S Structured Highlighter & 0.37 & 0.44 \\
    & H\&S Two Steps Highlighter & 0.35 & 0.44 \\
    & H\&S \deberta (\repliqa) & 0.27 & 0.22 \\
    & H\&S \deberta (\squad) & 0.22 & 0.16 \\
    & H\&S Span Highlighter & 0.19 & 0.18 \\
    \bottomrule
    \end{tabularx}
\end{table}

\autoref{tab:kprec-recall} presents K-precision and recall metrics for all pipelines across both datasets.
Vanilla RAG consistently achieves the highest K-precision scores (0.76 on RepliQA, 0.51 on BioASQ), outperforming all \hs variants by 6–10 percentage points.
K-precision measures the proportion of tokens in the predicted answer that appear in the reference passage, and these results suggest that Vanilla RAG tends to reproduce passage content more verbatim. In contrast, \hs pipelines produce answers that paraphrase or restructure the source material. The recall metrics show smaller gaps, suggesting that \hs pipelines still capture the essential information from reference answers despite lower lexical overlap.

A notable discrepancy emerges when comparing these token-overlap metrics with LLM judge evaluations (\autoref{fig:graders-wins} and \autoref{fig:llmjudges}.
This discrepancy reflects a fundamental distinction between lexical similarity and semantic correctness. K-precision rewards verbatim quotes from the source passage, which inflates scores for systems that extract and reproduce text directly. However, a correct answer does not need to share exact tokens with the reference: paraphrased responses that accurately convey the same information are equally valid.

\subsection{Decline to answer}

\begin{table}
\caption{Precision and Recall for declined answers.}
\label{tab:refuse}
\setlength{\tabcolsep}{5pt}
\begin{tabularx}{1.02\columnwidth}{Xrrr}
    \toprule
    Pipeline & Precision & Recall & F1 \\
    \midrule
    H\&S \deberta (\repliqa) & 0.85 & 0.99 & 0.91 \\
    H\&S Structured Highlighter & 0.94 & 0.40 & 0.57 \\
    H\&S Span Highlighter & 0.76 & 0.43 & 0.55 \\
    H\&S \deberta (\squad) & 0.55 & 0.48 & 0.51 \\
    Vanilla RAG & 0.94 & 0.24 & 0.38 \\
    H\&S Baseline & 0.60 & 0.20 & 0.30 \\
    H\&S Two Steps Highlighter & 0.45 & 0.01 & 0.03 \\
    \bottomrule
\end{tabularx}
\end{table}

Around 10\% of questions in the \repliqa dataset cannot be answered based on the provided document.
We evaluate how well the various pipelines declined to answer these questions.

In \autoref{tab:refuse}, we report K-Precision and Recall for each pipeline in terms of declining to answer.
We observe that \deberta-based models fare well, which is possibly due to the fact that their fine-tuning considers the no-answer case explicitly.
The \hs \deberta (\repliqa) performs particularly well; this may be partially due to the fact that it was trained on data with a similar distribution as the test set.
\hs Structured also does relatively well, possibly because the structured output option helps its reflection process.

Nevertheless, the absolute performance of all the pipelines suggests there is still significant potential for improvement in this aspect.
This connects to the open problem of teaching LLMs to say ``I don't know''~\cite{monteiro2024repliqa,madhusudhan2025llms,kirichenko2025abstentionbench}
Future \hs implementations should consider either better fine-tuning or few-shot prompt-tuning to improve on this aspect.

\subsection{Effect of \minwords}
\begin{figure}
    \centering
    \includegraphics[width=0.85\linewidth]{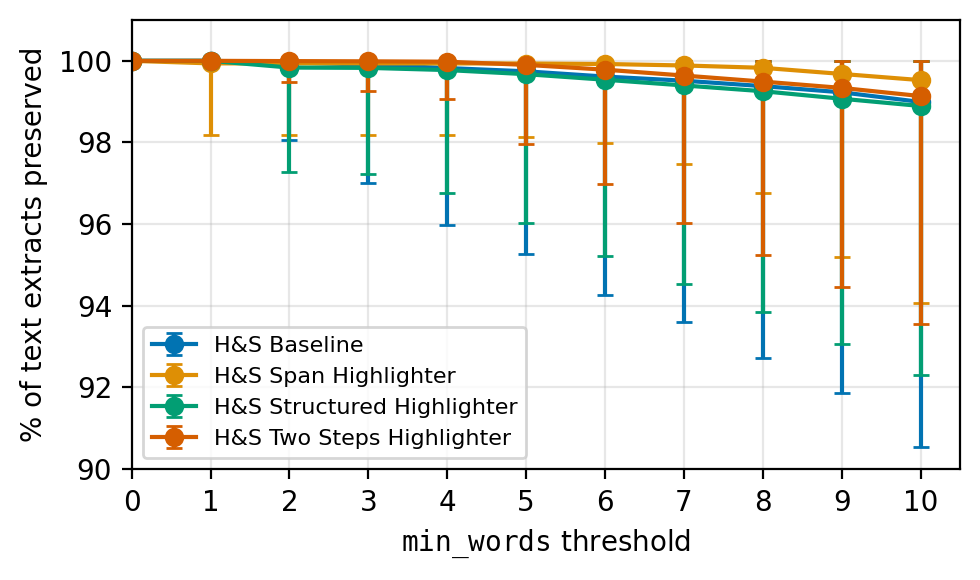}
    \caption{Effect of \minwords enforcement on the text extracts. It shows what percentage
    of text extracts satisfies the security property, for various levels of \minwords.
    Results on the \repliqa dataset.
    }
    \label{fig:hs-minwords}
\end{figure}

We measure the effect of enforcing the \minwords security parameters on the text extracts
from the various highlighter methods.
\autoref{fig:hs-minwords} shows that pipelines are only minimally affected by this:
the worst affected for $\minwords = 10$, a relatively large value for the security parameter,
is the \hs baseline with $99.0\% \pm 8.4$ loss of extracted texts.

\section{Ablation study of \hs components}
\label{sec:component_eval}

In this section we evaluate the individual components of an \hs pipeline, to provide deeper insights into \emph{why} this approach can achieve the results presented above.
First, we explore whether \hs is needed at all, or whether a simple passage retrieval or extraction \qa pipeline (i.e., highlighter without the summarizer) may suffice.
Next, we compare our different highlighter implementations in terms of their K-Precision and Recall.
Finally, we investigate whether the summarizer can recreate the original question based solely on the outputs of a highlighter.

\subsection{Do we need a generative summarizer?}
\label{sec:why-hs}

First, one may wonder whether we actually need the summarizer in the pipeline.
Given the long history of passage retrieval and extractive \qa, a simpler highlighter, which can be implemented on the basis of an extractive \qa model, may perform well in \qa tasks.

\begin{figure}
    \centering
    \includegraphics[width=1\linewidth]{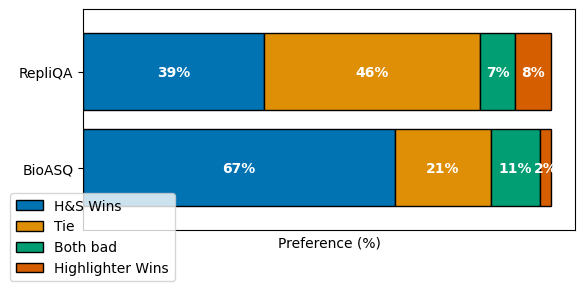}
    \caption{Do we need \hs, or can we just do highlighting
        (e.g., extractive \qa)?
        The preference is based on an LLM-based ComparisonJudge that looks specifically at the relevance and correctness of a response.}
    \label{fig:hs-vs-h}
\end{figure}

To answer this question, we compare the outputs of the individual highlighters against the response given by the full \hs pipeline using a ComparisonJudge.
To avoid longer answers from being judged more favorably, we truncate the responses to 40 words;
this is roughly 2 English sentences, which is what the \hs pipeline is prompted to produce.
We run this experiment for the Two Steps pipeline.

As shown in \autoref{fig:hs-vs-h}, our results indicate that using a generative LLM (summarizer) after the highlighting step can lead to substantial improvements.
This is particularly evident on the \bioasq dataset, where the \hs outputs were preferred in 67\% of cases versus only 2\% for the highlighter alone.
Based on a qualitative manual inspection of the ComparisonJudge's explanations accompanying its ratings, we observe that the \hs pipeline is typically preferred for its more ``relevant'' answers that ``directly address'' a question, whereas the highlighter is preferred for its ``direct quotes''.

\subsection{How good are the highlighters?}

We now compare the various highlighter implementations in terms of their K-Precision and Recall with respect to the gold passage.
Since this analysis requires a gold passage, we are only able to use the \repliqa dataset.

\begin{table}[h]
    \caption{Comparison between highlighter implementations with respect to the gold passage (\repliqa dataset.)}
    \label{tab:highlighter-comparison}
    \begin{tabularx}{1.02\columnwidth}{Xrr}
    \toprule
    Implementation & K-Precision & Recall \\
    \midrule
    H\&S Baseline & 0.85 & 0.66 \\
    H\&S Span Highlighter & 0.84 & 0.73 \\
    H\&S Structured Highlighter & 0.84 & 0.76 \\
    H\&S Two Steps Highlighter & 0.82 & 0.77 \\
    H\&S \deberta (\repliqa) & 0.80 & 0.36 \\
    H\&S \deberta (\squad) & 0.55 & 0.22 \\
    \bottomrule
    \end{tabularx}
\end{table}

As shown in \autoref{tab:highlighter-comparison}, we observe that LLM-based highlighters significantly outperform those based on \deberta, especially in terms of Recall.
However, as discussed in \autoref{sec:e2e-comparison}, LLM-based highlighters incur higher computational overheads, which may be a consideration in practical applications.
We did not focus on optimizing the performance of the \deberta models, and we hypothesize that there may still be scope for further improvement.
An encouraging result in this direction is that fine-tuning \deberta for the specific task of long-context highlighting (\repliqa dataset) improves performance compared to the same
model fine-tuned on \squad (short answers).
Results in \autoref{tab:highlighter-comparison} also indicate that using structured outputs for LLM-based highlighters improves performance (Recall).
However, as reported in \autoref{sec:e2e-comparison}, this incurs additional computational costs.

\begin{figure}
    \centering
    \includegraphics[width=1\linewidth]{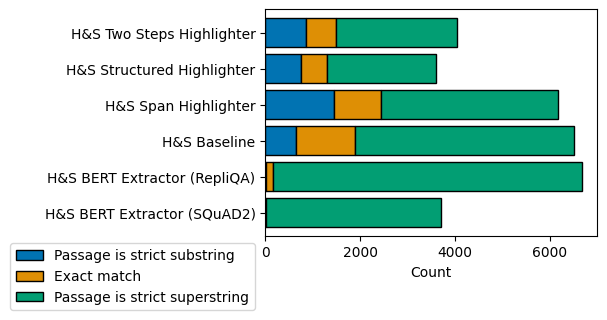}
    \caption{Comparison of the highlighter's output text to the human-curated gold passage in the \repliqa dataset.}
    \label{fig:highlighter-subsuperset}
\end{figure}

We further evaluate highlighters on a stricter metric: we count how many times the highlighter's output is either an exact match, a substring, or a superstring of the gold passage (although one should bear in mind that the span of the human-chosen gold passage is somewhat arbitrary).
\autoref{fig:highlighter-subsuperset} shows that, in this case, the \deberta highlighter that was fine-tuned on \repliqa has the best performance.
We attribute this to the fact that, even if its fine-tuning was done on a separate split of the \repliqa dataset, the model may have captured the style of gold passage selection of the human annotators for \repliqa.

\subsection{Can the summarizer guess the question?}

To gain a deeper understanding of the internal workings of an \hs pipeline, we investigate whether the summarizer can guess the user's original question, based solely on the text
provided by the highlighter.
We start by noting that this problem is, by its nature, ill-posed.
For example, consider the passage \textit{``In his later years, Kant lived a strictly ordered life. It was said that neighbors would set their clocks by his daily walks''}.\footnote{\url{https://en.wikipedia.org/wiki/Immanuel_Kant}.}
This text could plausibly answer several realistic questions, including \textit{``Was Immanuel Kant a creature of habit?''}, \textit{``What philosopher is best known for their punctuality?''}, or \textit{``Did Kant enjoy walking?''}.
Importantly, the passage would be a \textit{valid} answer to all those questions, implying that knowing the exact question is not strictly
necessary to answer.

\begin{table}[h]
    \caption{Can the summarizer guess the user's question? The summarizer guesses up to 10 questions, and we report
    the precision and recall for the best one of them.}
    \label{tab:guess-question}
    \begin{tabularx}{1.02\columnwidth}{p{1mm}Xrr}
    \toprule
    & Pipeline & K-Precision & Recall \\
    \midrule
    \multirow{6}{*}{\rotatebox[origin=c]{90}{\repliqa}}
    & \deberta (\repliqa) & 0.55 & 0.50 \\
    & Baseline & 0.55 & 0.56 \\
    & Structured Highlighter & 0.54 & 0.57 \\
    & Span Highlighter & 0.53 & 0.55 \\
    & Two Steps Highlighter & 0.52 & 0.57 \\
    & \deberta (\squad) & 0.37 & 0.38 \\
    \midrule
    \multirow{6}{*}{\rotatebox[origin=c]{90}{\bioasq}}
    & Span Highlighter & 0.70 & 0.46 \\
    & Structured Highlighter & 0.69 & 0.47 \\
    & Two Steps Highlighter & 0.69 & 0.47 \\
    & Baseline & 0.68 & 0.45 \\
    & \deberta (\repliqa) & 0.58 & 0.35 \\
    & \deberta (\squad) & 0.45 & 0.26 \\
    \bottomrule
    \end{tabularx}
\end{table}

When implementing the summarizer (\autoref{sec:hs}), we ask the LLM to output multiple \texttt{guessed\_questions}, a field that is not used by the \hs pipeline, but which helps with grounding the model.
\autoref{tab:guess-question} shows the K-Precision and Recall of the best among the \texttt{guessed\_questions} with respect to the true question.
As expected, the summarizer is rarely able to guess the question correctly.
However, its average performance is far from insignificant, with better pipelines achieving higher scores.
In \autoref{tab:guesses-repliqa_3-HSTwoStepsHighlighter_gpt_4_1_mini_gpt_4_1_mini}, we report some of the best and worst examples that highlight the ill-posedness of the problem.

These results confirm our intuition: it is hard to infer the question given just some highlighted context.
However, this also shows that guessing the right question is not necessary to answer a question, as evidenced by the great performance of our \hs pipelines \autoref{sec:e2e-comparison}.

\section{Discussion and future directions}
\label{sec:discussion}

This study of a new design pattern opens up a large number of research questions and directions to explore.
We discuss some of the challenges that we leave open for future work.

\paragraph{Automated checking for adversarial inputs in the knowledge base.}
Our threat model assumes a trustworthy knowledge base.
While this is realistic in many real-world use cases (e.g., \qa based on FAQs), it may not be the case in all RAG applications.
Interestingly, \hs makes it significantly easier to look for threats even in those cases: it (exponentially) reduces the scope of the problem from inspecting \textit{both} the knowledge base and the very large space of possible user inputs, down to inspecting only the knowledge base.
The system designer can scan their knowledge base for potential triggers, by inspecting each document via a (sequential) sliding window based of what the highlighter is allowed to highlight.
Future work can investigate efficient scanning techniques.

\begin{figure}
    \centering
    \includegraphics[width=0.3\textwidth]{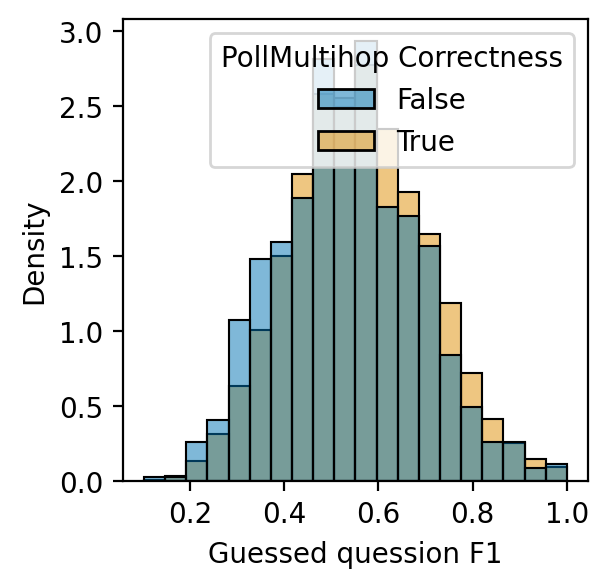}
    \caption{Weak correlation between the ability of the summarizer to guess the question
        and the performance of \hs Structured. Pearson correlation: 0.04 (p-value = $7 \times 10^{-7})$).
    }
    \label{fig:guessed-answer-vs-performance}
\end{figure}

\paragraph{Additional signals to the summarizer.}
Future work can study what additional information about the question can be passed from the highlighter to the summarizer without increasing
the risk of attacks.
For example, the \hs pipeline could monitor the similarity between the user's question and the question guessed by the summarizer. 
If there is too much divergence, the system could be programmed to refuse to answer, or to adapt the answer accordingly.
In preliminary experiments,
we found a weak correlation between the pipeline's performance and the summarizer's ability to predict the question: \autoref{fig:guessed-answer-vs-performance} shows the relationship between the full pipeline's performance and the K-Precision of the summarizer at guessing the question.

\paragraph{Handling \textit{Yes}/\textit{No} questions}
For some questions, a simple ``Yes'' or ``No'' can be a satisfactory answer.
We observe that \hs can be easily augmented to support such questions, whilst ensuring the same level of security.
For example, the highlighter can optionally pass a tag to the summarizer indicating:
i) whether the question being asked is of yes-no type, and if so
ii) what it thinks to be the answer (yes/no).
Since both of these fields are of boolean type, they are unlikely to introduce any new security risk.
The summarizer can then be instructed to augment its answer with this information.
Future work can investigate whether this approach improves answers for this class of questions.

\paragraph{Handling multiple questions.}
In some cases, the user may ask more than one question in the same prompt.
It may be possible to handle this via system design.
For example, the highlighter can first split the questions that are
present in the prompt, and then have the \hs system make separate calls to the summarizer.
Future work can investigate this design or other solutions to the problem.

\paragraph{Reasoning.}
We remark that a class of \qa tasks that \hs does not
support in its current stage is reasoning-based ones.
For example, it may be unable to answer questions such as
``Calculate the total number of holidays between annual leave and
national holidays'': in this case, the \hs pipeline will presumably
report quotes from both data sources (annual leave policy and
national holidays), but it will be unable to compute their sum,
as it cannot be aware that is the intent of the user.
We expect future research can hope to extend the applicability
of \hs to a wider set of tasks.
Nevertheless, we remark that \hs is readily useful for answering
matter of factual questions based on documents, such as policies
legal papers, or FAQ-style knowledge bases.

\paragraph{Does \hs reduce hallucinations?}
Intuitively, \hs should reduce hallucinations compared to standard RAG pipelines by sticking strictly to the documents retrieved from the knowledge base.
However, it is still possible that either the highlighter or summarizer could suffer from hallucinations.
For example, as discussed in \autoref{sec:e2e-comparison}, when faced with an unanswerable question, the highlighter sometimes highlighted sections that are either irrelevant or misleading.
Furthermore, the summarizer may suffer from hallucinations in its generative step, as usual.
Future work can investigate whether \hs, or some variant thereof, can help to reduce hallucinations.

\section{Conclusion}
We propose Highlight\&Summarize (\hs), a design pattern that enhances the generative step of a RAG pipeline to provide security by design against jailbreaking and model hijacking attacks.
We evaluate its security theoretically and empirically, against non-adaptive and adaptive adversaries.
Our empirical evaluations demonstrate that, compared to current RAG pipelines, our approach actually \emph{improves} the accuracy of responses in certain scenarios.
Overall, \hs provides strong security while enabling
high-performing RAG-style chatbots.
We also discuss several new research questions that arise from this design pattern, and encourage their study as exciting avenues for future research.

\section*{Acknowledgments}

We are grateful to Sahar Abdelnabi, Darren Edge, Adrien Ghosn, Jonathan Larson, and Santiago Zanella-Béguelin for helpful discussions.
We thank Elisa Alessandrini for help with revising a prior version of this manuscript.

\appendix

\section*{Open Science}

We share the following:
\begin{itemize}
    \item the code needed to run our experiments from scratch;
    \item the data artifacts the code produced, which include the responses from every question answering pipeline;
    \item a Jupyter notebook that enables parsing the data artifacts, and a clear reference to the ``cells'' needed to produce each of the tables and figures;
    \item \texttt{Lean 4} formalization of our theorems and their proofs;
    \item a demo application of H\&S (to be self-hosted) which answers questions based on our paper.
\end{itemize}

\paragraph{Accessing the artifacts.}
All artifacts are available at our
repository: \url{https://github.com/microsoft/highlight-summarize}.

\paragraph{Verifying our experimental results.}
To reproduce the tables and figures in this paper without re-running the experiments:
\begin{enumerate}
\item Navigate to the \texttt{reproduce/} folder.
\item The pre-computed results (pipeline outputs and LLM judge ratings) are stored under \texttt{reproduce/results/}. These files are tracked via Git LFS; if cloning the repository locally, run \texttt{git lfs pull --exclude=""} to download them.
\item Open the Jupyter notebook \texttt{reproduce/data-analysis.ipynb}. Each section is clearly labeled with the corresponding table or figure number from the paper (e.g., ``Table 3'', ``Figure 4'').
\end{enumerate}

\paragraph{Re-running experiments from scratch.}
To re-run all experiments, follow the setup instructions in the main \texttt{README.md}, configure Azure OpenAI authentication, then execute \texttt{python run\_experiments.py} from the \texttt{reproduce/} folder. The experiment configuration is specified in \texttt{experiments.yaml}.

\paragraph{Verifying the Lean 4 proofs.}
The formal proofs are located in \texttt{proofs/Proofs.lean}. To verify them, install Lean 4 (the required version is specified in \texttt{proofs/lean-toolchain}), then run \texttt{lake build} from the \texttt{proofs/} directory. Successful compilation confirms all theorems are proven.

\paragraph{Running the demo application.}
Install the demo dependencies with \texttt{pip install .[demo]}, then run \texttt{bash run.sh} from the \texttt{demo/} folder. The application will be accessible via a web browser.

\bibliographystyle{plain}
\bibliography{highlight-summarize}

\begin{thebibliography}{10}

\bibitem{abdelnabi2025llmail}
Sahar Abdelnabi, Aideen Fay, Ahmed Salem, Egor Zverev, Kai-Chieh Liao,
  Chi-Huang Liu, Chun-Chih Kuo, Jannis Weigend, Danyael Manlangit, Alex
  Apostolov, et~al.
\newblock {LLMail-Inject}: A dataset from a realistic adaptive prompt injection
  challenge.
\newblock {\em arXiv preprint arXiv:2506.09956}, 2025.

\bibitem{achiam2023gpt}
Josh Achiam, Steven Adler, Sandhini Agarwal, Lama Ahmad, Ilge Akkaya,
  Florencia~Leoni Aleman, Diogo Almeida, Janko Altenschmidt, Sam Altman,
  Shyamal Anadkat, et~al.
\newblock {GPT-4} technical report.
\newblock {\em arXiv preprint arXiv:2303.08774}, 2023.

\bibitem{adlakha2024evaluating}
Vaibhav Adlakha, Parishad BehnamGhader, Xing~Han Lu, Nicholas Meade, and Siva
  Reddy.
\newblock Evaluating correctness and faithfulness of instruction-following
  models for question answering.
\newblock {\em Transactions of the Association for Computational Linguistics},
  12:681--699, 2024.

\bibitem{alon2023detectingjailbreaks}
Gabriel Alon and Michael Kamfonas.
\newblock Detecting language model attacks with perplexity, 2023.

\bibitem{beurerkellner2025designpatterns}
Luca Beurer-Kellner, Beat Buesser, Ana-Maria Creţu, Edoardo Debenedetti,
  Daniel Dobos, Daniel Fabian, Marc Fischer, David Froelicher, Kathrin Grosse,
  Daniel Naeff, Ezinwanne Ozoani, Andrew Paverd, Florian Tramèr, and Václav
  Volhejn.
\newblock Design patterns for securing llm agents against prompt injections,
  2025.

\bibitem{bioasq}
{BioASQ}: A challenge on large-scale biomedical semantic indexing and question
  answering.
\newblock \url{https://www.bioasq.org}, 2012.
\newblock Accessed: 2024-05-02.

\bibitem{callan1994passage}
James~P Callan.
\newblock Passage-level evidence in document retrieval.
\newblock In {\em SIGIR’94: Proceedings of the Seventeenth Annual
  International ACM-SIGIR Conference on Research and Development in Information
  Retrieval, organised by Dublin City University}, pages 302--310. Springer,
  1994.

\bibitem{chen2019evaluating}
Anthony Chen, Gabriel Stanovsky, Sameer Singh, and Matt Gardner.
\newblock Evaluating question answering evaluation.
\newblock In {\em Proceedings of the 2nd workshop on machine reading for
  question answering}, pages 119--124, 2019.

\bibitem{chiang2023can}
Cheng-Han Chiang and Hung-yi Lee.
\newblock Can large language models be an alternative to human evaluations?
\newblock {\em arXiv preprint arXiv:2305.01937}, 2023.

\bibitem{chiang2024chatbot}
Wei-Lin Chiang, Lianmin Zheng, Ying Sheng, Anastasios~Nikolas Angelopoulos,
  Tianle Li, Dacheng Li, Banghua Zhu, Hao Zhang, Michael Jordan, Joseph~E
  Gonzalez, et~al.
\newblock Chatbot arena: An open platform for evaluating {LLMs} by human
  preference.
\newblock In {\em Forty-first International Conference on Machine Learning},
  2024.

\bibitem{costa2025fides}
Manuel Costa, Boris Köpf, Aashish Kolluri, Andrew Paverd, Mark Russinovich,
  Ahmed Salem, Shruti Tople, Lukas Wutschitz, and Santiago Zanella-Béguelin.
\newblock Securing ai agents with information-flow control, 2025.

\bibitem{fuzzyAI}
CyberArk.
\newblock Fuzzyai.
\newblock \url{https://github.com/cyberark/FuzzyAI}.

\bibitem{debenedetti2025caml}
Edoardo Debenedetti, Ilia Shumailov, Tianqi Fan, Jamie Hayes, Nicholas Carlini,
  Daniel Fabian, Christoph Kern, Chongyang Shi, Andreas Terzis, and Florian
  Tramèr.
\newblock Defeating prompt injections by design, 2025.

\bibitem{elo1967proposed}
Arpad~E Elo.
\newblock The proposed uscf rating system, its development, theory, and
  applications.
\newblock {\em Chess life}, 22(8):242--247, 1967.

\bibitem{hu2024tokenlevel}
Zhengmian Hu, Gang Wu, Saayan Mitra, Ruiyi Zhang, Tong Sun, Heng Huang, and
  Viswanathan Swaminathan.
\newblock Token-level adversarial prompt detection based on perplexity measures
  and contextual information, 2024.

\bibitem{jain2023baseline}
Neel Jain, Avi Schwarzschild, Yuxin Wen, Gowthami Somepalli, John Kirchenbauer,
  Ping yeh Chiang, Micah Goldblum, Aniruddha Saha, Jonas Geiping, and Tom
  Goldstein.
\newblock Baseline defenses for adversarial attacks against aligned language
  models, 2023.

\bibitem{jiang2024ragthief}
Changyue Jiang, Xudong Pan, Geng Hong, Chenfu Bao, and Min Yang.
\newblock Rag-thief: Scalable extraction of private data from
  retrieval-augmented generation applications with agent-based attacks, 2024.

\bibitem{jiang2006extraction}
Jing Jiang and Chengxiang Zhai.
\newblock Extraction of coherent relevant passages using hidden markov models.
\newblock {\em ACM Transactions on Information Systems (TOIS)}, 24(3):295--319,
  2006.

\bibitem{jm3}
Daniel Jurafsky and James~H. Martin.
\newblock {\em Speech and Language Processing: An Introduction to Natural
  Language Processing, Computational Linguistics, and Speech Recognition with
  Language Models}.
\newblock 3rd edition, 2025.
\newblock Online manuscript released January 12, 2025.

\bibitem{kirichenko2025abstentionbench}
Polina Kirichenko, Mark Ibrahim, Kamalika Chaudhuri, and Samuel~J Bell.
\newblock Abstentionbench: Reasoning {LLMs} fail on unanswerable questions.
\newblock {\em arXiv preprint arXiv:2506.09038}, 2025.

\bibitem{kumar2025certifying}
Aounon Kumar, Chirag Agarwal, Suraj Srinivas, Aaron~Jiaxun Li, Soheil Feizi,
  and Himabindu Lakkaraju.
\newblock Certifying {LLM} safety against adversarial prompting, 2025.

\bibitem{lee2016learning}
Kenton Lee, Shimi Salant, Tom Kwiatkowski, Ankur Parikh, Dipanjan Das, and
  Jonathan Berant.
\newblock Learning recurrent span representations for extractive question
  answering.
\newblock {\em arXiv preprint arXiv:1611.01436}, 2016.

\bibitem{madhusudhan2025llms}
Nishanth Madhusudhan, Sathwik~Tejaswi Madhusudhan, Vikas Yadav, and Masoud
  Hashemi.
\newblock Do {LLMs} know when to not answer? investigating abstention abilities
  of large language models.
\newblock In {\em Proceedings of the 31st International Conference on
  Computational Linguistics}, pages 9329--9345, 2025.

\bibitem{Microsoft2025SafetySystemMessages}
Microsoft.
\newblock Safety system messages, 2025.

\bibitem{monteiro2024repliqa}
Joao Monteiro, Pierre-Andre Noel, Etienne Marcotte, Sai~Rajeswar Mudumba,
  Valentina Zantedeschi, David Vazquez, Nicolas Chapados, Chris Pal, and Perouz
  Taslakian.
\newblock {RepliQA}: A question-answering dataset for benchmarking {LLMs} on
  unseen reference content.
\newblock {\em Advances in Neural Information Processing Systems},
  37:24242--24276, 2024.

\bibitem{mu2025closerlook}
Norman Mu, Jonathan Lu, Michael Lavery, and David Wagner.
\newblock A closer look at system prompt robustness, 2025.

\bibitem{o1975retrieval}
John O'Connor.
\newblock Retrieval of answer-sentences and answer-figures from papers by text
  searching.
\newblock {\em Information Processing \& Management}, 11(5-7):155--164, 1975.

\bibitem{o1980answer}
John O'Connor.
\newblock Answer-passage retrieval by text searching.
\newblock {\em Journal of the American Society for Information Science},
  31(4):227--239, 1980.

\bibitem{oosterhuis2024reliable}
Harrie Oosterhuis, Rolf Jagerman, Zhen Qin, Xuanhui Wang, and Michael
  Bendersky.
\newblock Reliable confidence intervals for information retrieval evaluation
  using generative ai.
\newblock In {\em Proceedings of the 30th ACM SIGKDD Conference on Knowledge
  Discovery and Data Mining}, 2024.

\bibitem{OpenAIModeration}
OpenAI.
\newblock Moderation.

\bibitem{pearce2021comparative}
Kate Pearce, Tiffany Zhan, Aneesh Komanduri, and Justin Zhan.
\newblock A comparative study of transformer-based language models on
  extractive question answering.
\newblock {\em arXiv preprint arXiv:2110.03142}, 2021.

\bibitem{peng2025dataextraction}
Yuefeng Peng, Junda Wang, Hong Yu, and Amir Houmansadr.
\newblock Data extraction attacks in retrieval-augmented generation via
  backdoors, 2025.

\bibitem{pisano2024bergeron}
Matthew Pisano, Peter Ly, Abraham Sanders, Bingsheng Yao, Dakuo Wang, Tomek
  Strzalkowski, and Mei Si.
\newblock Bergeron: Combating adversarial attacks through a conscience-based
  alignment framework, 2024.

\bibitem{prasad2023meetingqa}
Archiki Prasad, Trung Bui, Seunghyun Yoon, Hanieh Deilamsalehy, Franck
  Dernoncourt, and Mohit Bansal.
\newblock Meetingqa: Extractive question-answering on meeting transcripts.
\newblock In {\em Proceedings of the 61st Annual Meeting of the Association for
  Computational Linguistics (Volume 1: Long Papers)}, pages 15000--15025, 2023.

\bibitem{rajpurkar-etal-2016-squad}
Pranav Rajpurkar, Jian Zhang, Konstantin Lopyrev, and Percy Liang.
\newblock {SQ}u{AD}: 100,000+ questions for machine comprehension of text.
\newblock In Jian Su, Kevin Duh, and Xavier Carreras, editors, {\em Proceedings
  of the 2016 Conference on Empirical Methods in Natural Language Processing},
  pages 2383--2392, Austin, Texas, November 2016. Association for Computational
  Linguistics.

\bibitem{rebedea2023nemo}
Traian Rebedea, Razvan Dinu, Makesh Sreedhar, Christopher Parisien, and
  Jonathan Cohen.
\newblock Nemo guardrails: A toolkit for controllable and safe {LLM}
  applications with programmable rails, 2023.

\bibitem{robey2024smoothllm}
Alexander Robey, Eric Wong, Hamed Hassani, and George~J. Pappas.
\newblock {SmoothLLM}: Defending large language models against jailbreaking
  attacks, 2024.

\bibitem{operationBizarreBazaar}
Pillar Security.
\newblock Operation bizarre bazaar.
\newblock \url{https://www.pillar.security/resources/operation-bizarre-bazaar}.

\bibitem{shacham2007geometry}
Hovav Shacham.
\newblock The geometry of innocent flesh on the bone: return-into-libc without
  function calls (on the x86).
\newblock In {\em Proceedings of the 14th ACM Conference on Computer and
  Communications Security}, CCS '07, 2007.

\bibitem{verga2024replacing}
Pat Verga, Sebastian Hofstatter, Sophia Althammer, Yixuan Su, Aleksandra
  Piktus, Arkady Arkhangorodsky, Minjie Xu, Naomi White, and Patrick Lewis.
\newblock Replacing judges with juries: Evaluating {LLM} generations with a
  panel of diverse models.
\newblock {\em arXiv preprint arXiv:2404.18796}, 2024.

\bibitem{wang2022survey}
Luqi Wang, Kaiwen Zheng, Liyin Qian, and Sheng Li.
\newblock A survey of extractive question answering.
\newblock In {\em 2022 International Conference on High Performance Big Data
  and Intelligent Systems (HDIS)}, pages 147--153. IEEE, 2022.

\bibitem{xiang2024certifiably}
Chong Xiang, Tong Wu, Zexuan Zhong, David Wagner, Danqi Chen, and Prateek
  Mittal.
\newblock Certifiably robust {RAG} against retrieval corruption, 2024.

\bibitem{zhang2025corruptrag}
Baolei Zhang, Yuxi Chen, Minghong Fang, Zhuqing Liu, Lihai Nie, Tong Li, and
  Zheli Liu.
\newblock Practical poisoning attacks against retrieval-augmented generation,
  2025.

\bibitem{zheng2023judging}
Lianmin Zheng, Wei-Lin Chiang, Ying Sheng, Siyuan Zhuang, Zhanghao Wu, Yonghao
  Zhuang, Zi~Lin, Zhuohan Li, Dacheng Li, Eric Xing, et~al.
\newblock Judging {LLM}-as-a-judge with mt-bench and chatbot arena.
\newblock {\em Advances in Neural Information Processing Systems}, 2023.

\bibitem{zou2024poisonedrag}
Wei Zou, Runpeng Geng, Binghui Wang, and Jinyuan Jia.
\newblock {PoisonedRAG}: Knowledge corruption attacks to retrieval-augmented
  generation of large language models, 2024.

\end{thebibliography}
\balance

\cleardoublepage

\section{Proofs}

\bounded*
\begin{proof}
    For a fixed input $p$, the control region is $\mathcal{C}_{\beta}^p(\llm) = \{o \in \llmoutput \mid P(\llm(p) \mid o) \geq \beta\}$.
    Because $\sum_o P(\llm(p) = o) = 1$, there can be at most $\frac{1}{\beta}$
    outputs with probability $\geq \beta$.

    Then
    $$|\mathcal{C}_{\beta}(\llm)| \leq \left|\bigcup_{p\in \llminput} \mathcal{C}_{\beta}^p(\llm)\right| \leq \sum_{p\in \llminput} |\mathcal{C}_{\beta}^p(\llm)| \leq \frac{|\llminput|}{\beta} \,.
    $$
\end{proof}

\exponential*

\begin{proof}

We observe that \hs limits the input space to the set of (contiguous) substrings of the document
that have at most length $K$.
Hence, by \autoref{thm:bounded-control-region},
$|\mathcal{C}_{\beta}(\llm \circ h_D)| \leq \nicefrac{K N}{\beta}$.

Remember that $\llminput = \Sigma^{\leq K}$, hence $|\llminput| = O(|\Sigma|^K)$.
The result stems from applying the assumption $|\mathcal{C}_{\beta}(\llm)| \geq \alpha |\llminput|$
to obtain:
$$\frac{|\mathcal{C}_{\beta}(\llm \circ h_D)|}{|\mathcal{C}_{\beta}(\llm)|} \leq \frac{K N}{\alpha\beta|\llminput|} \,.$$

\end{proof}

\section{\hs costs}
\label{sec:costs}

\paragraph{Time.}
We measure the time taken by each pipeline to generate an answer, averaged over 40 questions, as shown in \autoref{tab:timing}.
The absolute time values are specific to our setup, but the relative differences show that, as expected, the LLM-based \hs pipelines take longer to generate an answer.
We note that our implementation has not yet been optimized for performance,
and we expect future work can improve these overheads.

\begin{figure}
    \centering
    \includegraphics[width=0.55\linewidth]{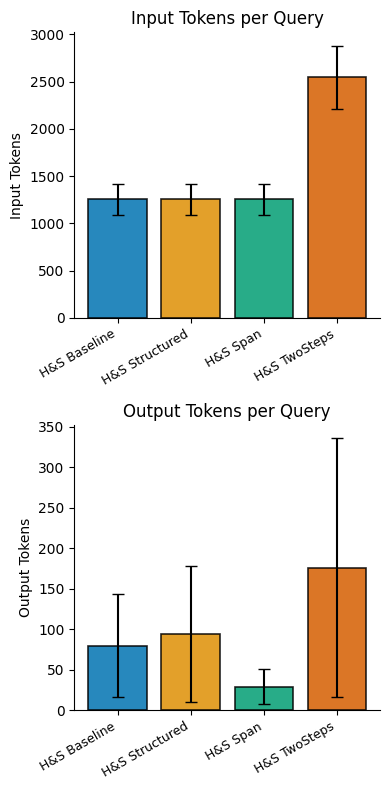}
    \caption{Average output tokens for the different highlighters.}
    \label{fig:hs-costs}
\end{figure}

\paragraph{Tokens.}
We measure the number of output tokens for the various pipelines (\autoref{fig:hs-costs}).
Unsurprisingly, we observe that Two Steps has the highest costs (roughly twice that of the other pipelines), followed by the Structured highlighter.
\hs Span is a promising avenue, as it reduces both computational and token
costs, but its performance lacks and needs to be improved in future work.

\begin{table}
\centering
\caption{Processing time for one question, averaged over 40 queries on the \repliqa dataset.}
\label{tab:timing}
\begin{tabular}{lr}
\toprule
Pipeline & Time (s) \\
\midrule
Vanilla RAG & 0.78 \\
H\&S \deberta (\repliqa) & 1.68 \\
H\&S \deberta (\squad) & 2.00 \\
H\&S Span Highlighter & 4.19 \\
H\&S Baseline & 4.62 \\
H\&S Structured Highlighter & 5.89 \\
H\&S Two Steps Highlighter & 8.66 \\
\bottomrule
\end{tabular}
\end{table}

\section{Additional results}

We report the full results from the 1-to-1 comparison between pipelines in \autoref{tab:battle_results},
and the question guesses of \hs Two Steps on \repliqa
\autoref{tab:guesses-repliqa_3-HSTwoStepsHighlighter_gpt_4_1_mini_gpt_4_1_mini}.

\begin{table*}
\scriptsize
\caption{Pairwise battle results: Each cell shows Wins (W) and Ties (T) for the row model vs the column model.}
\label{tab:battle_results}
\scriptsize
\begin{tabular}{llrrrrrrrrrrrrrrrr}
\toprule
 &  & \multicolumn{2}{c}{Span} & \multicolumn{2}{c}{Baseline} & \multicolumn{2}{c}{BERT S.} & \multicolumn{2}{c}{BERT R.} & \multicolumn{2}{c}{Two Steps} & \multicolumn{2}{c}{Structured} & \multicolumn{2}{c}{Vanilla} & \multicolumn{2}{c}{Total} \\
\cmidrule(lr){3-16} \cmidrule(lr){17-18}
 &  & W & T & W & T & W & T & W & T & W & T & W & T & W & T & W & T \\
\midrule
\multirow{7}{*}{\rotatebox[origin=c]{90}{\repliqa}}
& Span & -- & -- & 3563 & 10349 & 14669 & 2565 & 12718 & 4354 & 1480 & 12039 & 2266 & 11768 & 5702 & 7786 & 40398 & 48861 \\
& Baseline & 4043 & 10349 & -- & -- & 15310 & 2207 & 13342 & 3883 & 1521 & 12496 & 2770 & 11276 & 5777 & 7937 & 42763 & 48148 \\
& BERT (S.) & 721 & 2565 & 438 & 2207 & -- & -- & 2955 & 8742 & 210 & 1701 & 428 & 2248 & 354 & 1814 & 5106 & 19277 \\
& BERT (R.) & 883 & 4354 & 730 & 3883 & 6258 & 8742 & -- & -- & 502 & 3409 & 429 & 4022 & 1189 & 3391 & 9991 & 27801 \\
& Two Steps & 4436 & 12039 & 3938 & 12496 & 16044 & 1701 & 14044 & 3409 & -- & -- & 3038 & 13607 & 4967 & 9079 & 46467 & 52331 \\
& Structured & 3921 & 11768 & 3909 & 11276 & 15279 & 2248 & 13504 & 4022 & 1310 & 13607 & -- & -- & 3982 & 9974 & 41905 & 52895 \\
& Vanilla & 4467 & 7786 & 4241 & 7937 & 15787 & 1814 & 13375 & 3391 & 3909 & 9079 & 3999 & 9974 & -- & -- & 45778 & 39981 \\
\midrule
\multirow{7}{*}{\rotatebox[origin=c]{90}{\bioasq}}
& Span & -- & -- & 468 & 1075 & 1815 & 1857 & 1542 & 1599 & 276 & 1166 & 344 & 1024 & 544 & 763 & 4989 & 7484 \\
& Baseline & 3176 & 1075 & -- & -- & 3894 & 634 & 3494 & 882 & 675 & 2834 & 871 & 2506 & 1928 & 1372 & 14038 & 9303 \\
& BERT (S.) & 1047 & 1857 & 191 & 634 & -- & -- & 820 & 2094 & 99 & 531 & 113 & 520 & 111 & 318 & 2381 & 5954 \\
& BERT (R.) & 1578 & 1599 & 343 & 882 & 1805 & 2094 & -- & -- & 217 & 910 & 206 & 883 & 379 & 682 & 4528 & 7050 \\
& Two Steps & 3277 & 1166 & 1210 & 2834 & 4089 & 531 & 3592 & 910 & -- & -- & 867 & 3240 & 1614 & 1566 & 14649 & 10247 \\
& Structured & 3351 & 1024 & 1342 & 2506 & 4086 & 520 & 3630 & 883 & 612 & 3240 & -- & -- & 2060 & 1301 & 15081 & 9474 \\
& Vanilla & 3412 & 763 & 1419 & 1372 & 4290 & 318 & 3658 & 682 & 1539 & 1566 & 1358 & 1301 & -- & -- & 15676 & 6002 \\
\bottomrule
\end{tabular}
\end{table*}

\begin{table*}
\caption{Best and worst question guesses by the summarizer of the \hs Two Steps Highlighter pipeline on RepliQA.}
\label{tab:guesses-repliqa_3-HSTwoStepsHighlighter_gpt_4_1_mini_gpt_4_1_mini}
\begin{tabularx}{\linewidth}{cXX}
\toprule
Score & Real question & Guessed question \\
\midrule
1.00 & What event led to the creation of the Smithville Fiscal Responsibility Committee? & What led to the creation of the Smithville Fiscal Responsibility Committee? \\
1.00 & What new technologies did Chen introduce to Wanderlust Trails due to the COVID-19 pandemic? & What technologies did Chen introduce? \\
1.00 & How did the Melville Bay small businesses respond to the implementation of the Melville Cybersecurity Compliance Act (MCCA) on November 3, 2023? & How did businesses respond to the implementation of the Melville Cybersecurity Compliance Act? \\
\midrule
0.18 & Which start-up surpassed its crowdfunding goal by April after its establishment in October 2023? & Why did ComfyTech choose crowdfunding for their project? \\
0.15 & Which renowned professor recently made a significant biblical discovery concerning giants? & What recent discoveries have been made about the Nephilim? \\
0.12 & Can you name the play that will premiere at Bensonville Community Playhouse on November 19, 2023? & When is 'The Whimsical World of Windermere' scheduled to premiere? \\
\bottomrule
\end{tabularx}
\end{table*}

\end{document}